\documentclass{article}

\usepackage{microtype}
\usepackage{graphicx}
\usepackage{subfigure}
\usepackage{booktabs} %

\usepackage{hyperref}

\usepackage[accepted]{icml2021}

\usepackage{times}
\usepackage{latexsym}

\usepackage[T1]{fontenc}

\usepackage{inconsolata}

\usepackage{xspace}
\usepackage{hyperref}       %
\usepackage{url}            %
\usepackage{booktabs}       %
\usepackage{amsfonts}       %
\usepackage{nicefrac}       %
\usepackage{microtype}      %
\usepackage{color}
\usepackage{multirow}
\usepackage{listings}

\usepackage{tabularx}
\usepackage{placeins}

\newcommand{\secref}[1]{\S\ref{#1}\xspace}
\newcommand{\DSP}{\textsc{DSP}\xspace}

\newcommand{\demonstrate}{\textsc{Demonstrate}\xspace}
\newcommand{\search}{\textsc{Search}\xspace}
\newcommand{\predict}{\textsc{Predict}\xspace}

\definecolor{ourdarkblue}{HTML}{0499CC}
\definecolor{ourlightblue}{HTML}{03A9F4}
\definecolor{ourdarkgray}{HTML}{838A8A}
\definecolor{ourlightgray}{HTML}{B8B8B8}
\definecolor{ourgreen}{HTML}{4D8951}
\definecolor{ourblack}{HTML}{212121}
\definecolor{oursteelblue}{HTML}{9BB8D7}
\definecolor{ourorange}{HTML}{FDBA58}
\definecolor{ourwhite}{HTML}{FAFAFA}
\definecolor{ourpurple}{HTML}{876DB5}
\definecolor{ourmaroon}{HTML}{881C1c}
\definecolor{superlightgray}{HTML}{DDDDDD}

\newcommand{\lmcolor}[1]{#1} %
\newcommand{\ircolor}[1]{#1} %

\newcommand{\LM}{{\color{ourdarkblue}\lmcolor{\textrm{\textbf{LM}}}}}
\newcommand{\RM}{{\color{ourgreen}\ircolor{\textrm{\textbf{RM}}}}}

\definecolor{codegreen}{rgb}{0,0.6,0}
\definecolor{codegray}{rgb}{0.5,0.5,0.5}

\definecolor{backcolour}{RGB}{245,248,250}
\definecolor{emph}{RGB}{166,88,53}
\definecolor{nightblue}{RGB}{9,49,105}
\definecolor{keywords}{RGB}{207,33,46}
\definecolor{lightpurple}{RGB}{130,81,223}

\lstdefinestyle{mystyle}{
    backgroundcolor=\color{backcolour},   
    commentstyle=\color{codegreen},
    keywordstyle=\color{keywords},
    stringstyle=\color{nightblue},
    basicstyle=\fontsize{7}{8}\ttfamily,
    breakatwhitespace=true,         
    breaklines=true,                 
    captionpos=b,                    
    keepspaces=true,                 
    numberstyle=\tiny\color{codegray},
    numbersep=2pt,                  
    showspaces=false,                
    showstringspaces=false,
    showtabs=false,                  
    tabsize=2,
    emph={dsp,Example,sample,annotate,knn,crossval,generate,retrieve,retrieve\_ensemble,majority,fused_retrieval,Template, Transformation,rank,branch},
    emphstyle={\color{lightpurple}},
    linewidth=0.98\columnwidth,
    frame=tb,    
    xrightmargin=0pt,
    xleftmargin=0.23cm,
    numbers=left,
    aboveskip=0.4cm,
    belowskip=0.4cm,
}

\icmltitlerunning{\demonstrate{}--\search{}--\predict{}: Composing retrieval and language models}

\begin{document}

\twocolumn[
\icmltitle{\demonstrate{}--\search{}--\predict{}: \\ Composing retrieval and language models for knowledge-intensive NLP}

\icmlsetsymbol{equal}{*}

\begin{icmlauthorlist}
\icmlauthor{Omar Khattab}{stanford}\hspace{5mm}
\icmlauthor{Keshav Santhanam}{stanford}\hspace{5mm}
\icmlauthor{Xiang Lisa Li}{stanford}\hspace{5mm}
\icmlauthor{David Hall}{stanford}
\\
\icmlauthor{Percy Liang}{stanford}\hspace{5mm}
\icmlauthor{Christopher Potts}{stanford}\hspace{5mm}
\icmlauthor{Matei Zaharia}{stanford}
\end{icmlauthorlist}

\icmlaffiliation{stanford}{\textbf{Stanford University}}

\icmlcorrespondingauthor{\\ \textbf{Omar Khattab}}{\texttt{okhattab@cs.stanford.edu}}

\icmlkeywords{Machine Learning, ICML}

\vskip 0.3in
]

\printAffiliationsAndNotice{}  %

\begin{abstract}
Retrieval-augmented in-context learning has emerged as a powerful approach for addressing knowledge-intensive tasks using frozen language models (LM) and retrieval models (RM).
Existing work has combined these in simple ``retrieve-then-read'' pipelines in which the RM retrieves passages that are inserted into the LM prompt.
To begin to fully realize the potential of frozen LMs and RMs, we propose \demonstrate--\search--\predict (\DSP), a framework that relies on passing natural language texts in sophisticated pipelines between an LM and an RM.
\DSP can express high-level programs that bootstrap pipeline-aware demonstrations, search for relevant passages, and generate grounded predictions, systematically breaking down problems into small transformations that the LM and RM can handle more reliably.
We have written novel \DSP programs %
for answering questions in open-domain, multi-hop, and conversational settings, establishing in early evaluations new state-of-the-art in-context learning results and delivering 37--120\%, 8--39\%, and 80--290\% relative gains against the vanilla LM (GPT-3.5), a standard retrieve-then-read pipeline, and a contemporaneous self-ask pipeline, respectively. We release \DSP at \url{https://github.com/stanfordnlp/dsp}.
\end{abstract}

\section{Introduction}

In-context learning adapts a frozen language model (LM) to tasks by conditioning the LM on a textual prompt including task instructions and a few demonstrating examples~\cite{mccann2018natural,radford2019language,brown2020language}. For knowledge-intensive tasks such as question answering, fact checking, and information-seeking dialogue, retrieval models (RM) are increasingly used to augment prompts with relevant information from a large corpus~\citep{lazaridou2022internet,press2022measuring,khot2022decomposed}.

Recent work has shown such \textit{retrieval-augmented in-context learning} to be effective in simple ``retrieve-then-read'' pipelines: a query is fed to the RM and the retrieved passages become part of a prompt that provides context for the LM to use in its response. In this work, we argue that the fact that both LMs and RMs consume (and generate or retrieve) natural language texts creates an opportunity for much more sophisticated interactions between them. Fully realizing this would be transformative: frozen LMs and RMs could serve as infrastructure across tasks, enabling ML- and domain-experts alike to rapidly build grounded AI systems at a high level of abstraction and with lower deployment overheads and annotation costs.

\begin{figure}[!t]
    \centering
    \includegraphics[width=.95\columnwidth]{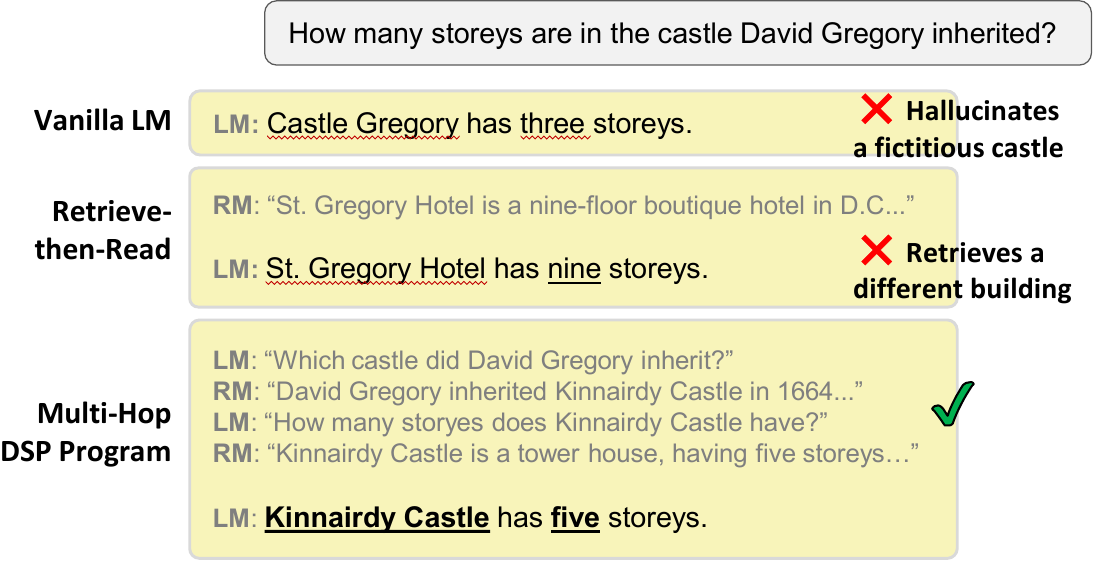}
    \vspace{-3mm}
    \caption{A comparison between three systems based on GPT-3.5 (\texttt{text-davinci-002}). On its own, the LM often makes false assertions. An increasingly popular retrieve-then-read pipeline fails when simple search can't find an answer. In contrast, a task-aware DSP program successfully decomposes the problem and produces a correct response. Texts edited for presentation.}
    \label{fig:tasks}
    \vspace{-6mm}
\end{figure}

\begin{figure*}[!t]
    \centering
    \includegraphics[width=.96\textwidth]{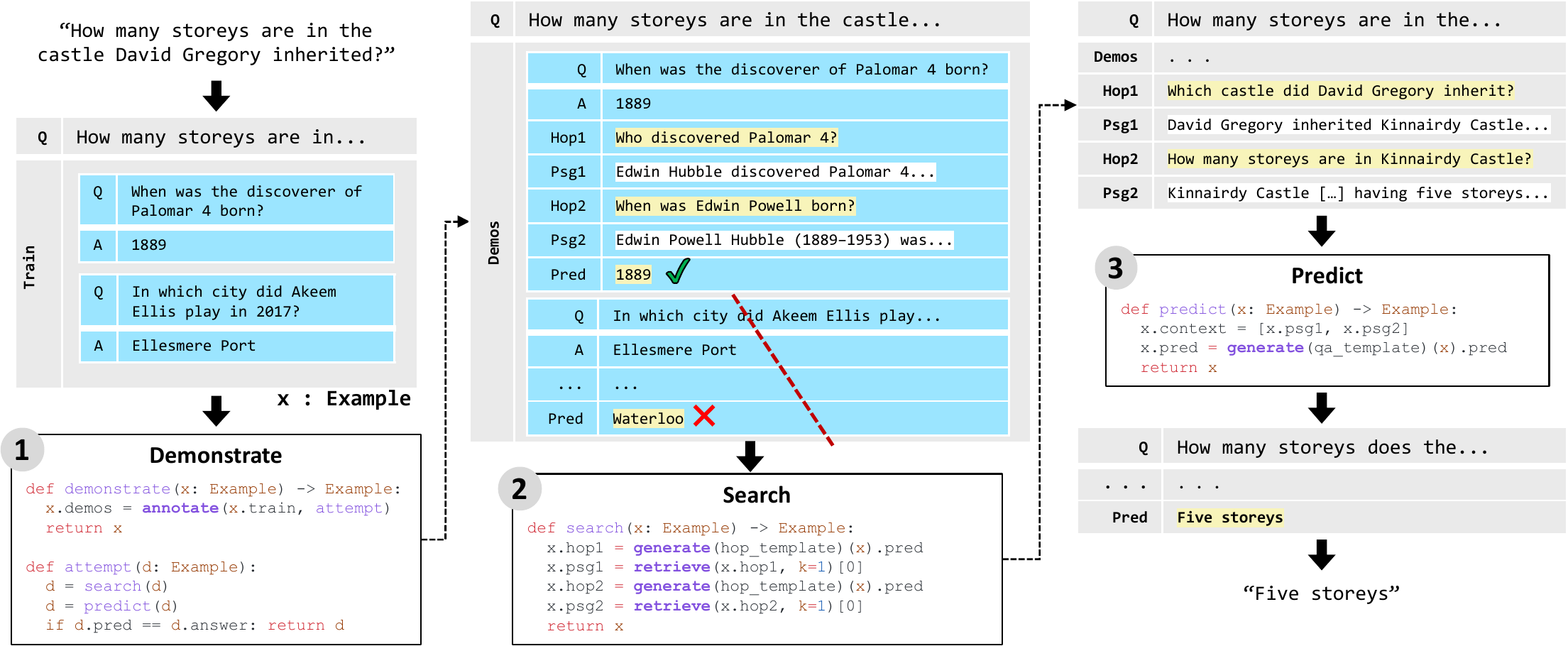}
    \vspace{-2mm}
    \caption{A toy example of a DSP program for multi-hop question answering. Given an input question and a 2-shot training set, the \demonstrate stage programmatically annotates intermediate transformations on the training examples using a form of weak supervision. Learning from a resulting \textit{demonstration}, the \search stage decomposes the complex input question and retrieves supporting information over two retrieval hops. Finally, the \predict stage uses the demonstration and retrieved passages to answer the question.}
    \label{fig:example}
    \vspace{-3mm}
\end{figure*}

Figure~\ref{fig:tasks} begins to illustrate the power of retrieval-augmented in-context learning, but also the limitations of ``retrieve-then-read'' \cite{lazaridou2022internet,izacard2022few}. Our query is ``How many storeys are in the castle David Gregory inherited?'' When prompted to answer this, GPT-3.5 (\texttt{text-davinci-002}; \citealt{ouyang2022training}) makes up a fictitious castle with incorrect attributes, highlighting the common observation that knowledge stored in LM parameters is often unreliable~\cite{shuster2021retrieval,ishii2022survey}.
Introducing an \RM{} component helps, as the \LM{} can ground its responses in retrieved passages, but a rigid retrieve-then-read strategy fails because the \RM{} cannot find passages that directly answer the question.

We introduce the \textbf{\demonstrate--\search--\predict} (\textbf{\DSP}) framework for in-context learning, which relies entirely on passing natural language text (and scores) between a frozen \RM{} and \LM{}. \DSP introduces a number of composable functions that bootstrap training examples (\demonstrate), gather information from a knowledge corpus (\search), and generate grounded outputs (\predict), using them to systematically unify techniques from the retrieval-augmented NLP and the in-context learning literatures~\citep{lee2019latent,khattab2021baleen,anantha2020open,gao2022attributed,izacard2022few,dohan2022language,zelikman2022star,zhang2022automatic}. 
We use \DSP to suggest powerful %
strategies for knowledge-intensive tasks with compositions of these techniques. This reveals new conceptual possibilities for in-context learning in general (\secref{sec:dsp}), and it allows us to present rich programs that set new state-of-the-art results (\secref{sec:eval}).

Figure~\ref{fig:tasks} shows the path that a \DSP program might take to arrive at an answer, and Figure~\ref{fig:example} illustrates how a deliberate program achieves this. Instead of asking the \LM{} to answer this complex question, the program's \search stage uses the \LM{} to generate a query ``Which castle did David Gregory inherit?'' The \RM{} retrieves a passage saying Gregory inherited the Kinnairdy Castle. After a second search ``hop'' finds the castle's number of storeys, the \predict stage queries the \LM{} with these passages to answer the original question. Although this program implements behaviors such as query generation, it requires no hand-labeled examples of these intermediate \textit{transformations} (i.e., of the queries and passages of both retrieval hops). Instead, the \demonstrate stage uses labeled question--answer pairs to implement a form of weak supervision that programmatically annotates the transformations invoked within \search and \predict.

We evaluate several \DSP programs on answering questions in open-domain, multi-hop, and conversational settings. In them, we implement novel and reusable transformations such as bootstrapping annotations for all of our pipelines with weak supervision (\secref{sec:dsp:demonstrate}), reliably rewriting questions to resolve conversational dependencies and iteratively decompose complex queries with summarization of intermediate hops (\secref{sec:dsp:search}), and generating grounded responses from multiple passages with self-consistency (\secref{sec:dsp:predict}). We report preliminary results on Open-SQuAD, HotPotQA, and QReCC using the frozen \LM{} GPT-3.5 and \RM{} ColBERTv2~\cite{khattab2020colbert,santhanam-etal-2022-colbertv2} with no fine-tuning. Our \DSP programs deliver 37--120\%, 8--39\%, and 80--290\% relative gains against corresponding vanilla LMs, a standard retrieve-then-read pipeline, and a contemporaneous self-ask pipeline \citep{press2022measuring}, respectively.
Future versions of this report will include additional test tasks and \LM\ choices.

In summary, this work makes the following contributions. First, we argue that simple task-agnostic pipelines for in-context learning should give way to deliberate, task-aware strategies. Second, we show that this shift need not be a burden: with \DSP, such strategies can be easily expressed as short programs using composable operators. Third, this composability spawns powerful capacities, like automatically annotating demonstrations for complex pipelines from end-task labels. Fourth, for three knowledge-intensive tasks, we implement rich programs that establish state-of-the-art results for in-context learning.

\section{\demonstrate--\search--\predict} \label{section:dsp} \label{sec:dsp}

We now introduce the \DSP framework and show its expressive power by suggesting a number of strategies in which the \LM{} and \RM{} can come together to tackle complex problems effectively. We show in \secref{sec:eval} that such strategies outperform existing in-context learning methods. We begin by discussing the \LM{} and \RM{} foundation modules on which \DSP is built (\secref{sec:dsp:modules}) and then the datatypes and control flow within \DSP (\secref{sec:dsp:concepts}). Subsequently, we discuss each of the three inference stages: \demonstrate (\secref{sec:dsp:demonstrate}), \search (\secref{sec:dsp:search}), and \predict (\secref{sec:dsp:predict}).

\subsection{Pretrained Modules: \LM{} and \RM{}}
\label{sec:dsp:modules}

A \DSP program defines the communication between the language model \LM{} and the retrieval model \RM{}.%

\vspace{-2mm}
\paragraph{Language Model} We invoke a frozen language model \LM{} to conditionally \texttt{generate} (or \texttt{score}) text. For each invocation, the program prepares a \textit{prompt} that adapts the \LM{} to a specific function (e.g., answering questions or generating queries). A prompt often includes instructions, a few demonstrations of the desired behavior, and an input query to be answered.

As in Figure~\ref{fig:example}, the \LM{} generates not only: \textbf{(i)} the final answer to the input question (in the \predict stage), but also \textbf{(ii)} intermediate ``hop'' queries to find useful information for the input question (\search)  as well as \textbf{(iii)} exemplar queries that illustrate how to produce queries for questions in the training set (\demonstrate). This systematic use of the \LM{} is a hallmark of \DSP programs.

\vspace{-2mm}
\paragraph{Retrieval Model} \DSP programs also invoke a frozen retrieval model \RM{} to \texttt{retrieve} the top-$k$ most ``relevant'' text sequences for a given \textit{query}. The \RM{} can \texttt{index} a massive set of pre-defined \textit{passages} for scalable search, and those passages can be updated without changing the retrieval parameters. %
The \RM{} accepts free-form textual inputs and specializes in estimating the relevance (or similarity) of a text sequence to a query.

As in Figure~\ref{fig:example}, the \RM{} is responsible for retrieving \textbf{(i)} passages for each query generated by the \LM{} (during the \search stage), but also \textbf{(ii)} passages that are used within demonstrations (\demonstrate). In the latter case, the \RM's contributions are less about providing directly relevant information to the input question and more about helping the \LM{} adapt to the domain and task.

Though not utilized in this example, the \RM{} is also used in \DSP for functions like retrieving ``nearest-neighbor'' demonstrations from task training data (\demonstrate) and selecting well-grounded generated sequences from the \LM{} (\predict).

\subsection{Datatypes and Control Flow}
\label{sec:dsp:concepts}

We have implemented the \DSP{} framework in Python. The present section introduces the core data types and composable functions provided by the framework. We use illustrative code snippets to ground the examples, and to convey the power that comes from being able to express complex interactions between the \LM{} and \RM{} in simple programs.

\vspace{-2mm}
\paragraph{The \texttt{Example} Datatype} To conduct a task, a \DSP program manipulates one or more instances of the \texttt{Example} datatype. An \texttt{Example} behaves like a Python dictionary with multiple fields. The program is typically provided with a few training examples. The code snippet below illustrates this for multi-hop question answering.

\begin{samepage}
\begin{lstlisting}[language=Python,breaklines=true,showstringspaces=false]
from dsp import Example

train = [Example(question="When was the discoverer of Palomar 4 born?", answer="1889"),
         Example(question="In which city did Akeem Ellis play in 2017?", answer="Ellesmere Port")]
\end{lstlisting}
\end{samepage}

This snippet contains two labeled examples, each with a multi-hop question (e.g., ``In which city did Akeem Ellis play in 2017?'')\ and its short answer (``Ellesmere Port''). Arbitrary keys and values are allowed within an \texttt{Example}, though typical values are strings or lists of strings. 

In this task, we are unlikely to find an individual passage that provides the answer to any question. For example, the first training example can probably be resolved only by first answering the question of who discovered Palomar (``Edwin Hubble'') and then addressing the question of Hubble's birth date using different evidence passages. We typically assume that the human-labeled training data do \textit{not} include labels for intermediate transformations (e.g., queries for individual hops) that would be useful for following these steps, and so it is the job of the \DSP{} program to discover these strategies via in-context learning. %

\vspace{-2mm}
\paragraph{A \DSP Program} The following code snippet is a complete program for resolving multi-hop questions like those in Figure~\ref{fig:tasks}, with help from train examples like those above.

\begin{samepage}
\begin{lstlisting}[language=Python,breaklines=true,showstringspaces=false]
def multihop_program(question: str) -> str:
   x = Example(question=question, train=train)
   x = multihop_demonstrate(x)
   x = multihop_search(x)
   x = multihop_predict(x)
   return x.answer

multihop_program("How many storeys does the castle David Gregory inherited have?")
# => "five storeys"
\end{lstlisting}
\end{samepage}

The program takes the input (here, a question) and outputs the system output (its short answer). It starts by creating an \texttt{Example} for the input question and assigning the \texttt{train} field to the training set from the previous snippet. Programs invoke and compose \DSP \textit{primitives} (i.e., built-in functions) to build the \demonstrate, \search, and \predict transformations that define the program.

\vspace{-2mm}
\paragraph{Transformations} A transformation is a function that takes an \texttt{Example} as input and returns an  \texttt{Example}, populating new fields (or modifying existing fields) in it. 
This program invokes three developer-defined transformations, namely, \texttt{multihop\_demonstrate}, \texttt{multihop\_search}, and \texttt{multihop\_predict}. Transformations may themselves invoke other transformations, and they act analogously to layers in standard deep neural network (DNN) programming frameworks such as PyTorch, except that they pass text data instead of tensors between each other and do not involve backpropagation.

We categorize transformations according to their behavior (or purpose) under one of the \demonstrate, \search, and \predict stages. That said, \DSP does not impose this categorization and allows us to define functions that may blend these stages. We will discuss each of the three stages next.

\subsection{\demonstrate}
\label{sec:dsp:demonstrate}

 It is known that including examples of the desired behavior from the \LM{} in its prompt typically leads to better performance \cite{brown2020language}. In \DSP, a \textit{demonstration} is a training example that has been prepared to illustrate specific desired behaviors from the \LM{}. A \demonstrate transformation takes as input \texttt{x} of type \texttt{Example} and prepares a list of demonstrations in \texttt{x.demos}, typically by \textit{selecting} a subset of the training examples in \texttt{x.train} and \textit{bootstrapping} new fields in them.

\vspace{-2mm}
\paragraph{Bootstrapping Demonstrations} Examples in the training set typically consist of the input text and the target output of the task. The \demonstrate stage can augment a training example by programmatically bootstrapping annotations for intermediate transformations. In our running ``multi-hop'' example, the demonstrations illustrate three \LM{}-based transformations: \textbf{(i)} how to break down the input question in order to gather information for answering it (i.e., first-hop retrieval), \textbf{(ii)} how to use information gathered in an earlier ``hop'' to ask follow-up questions, and \textbf{(iii)} how to use the information gathered to answer complex questions.

\begin{samepage}
\begin{lstlisting}[language=Python,breaklines=true,showstringspaces=false]
Examples = list[Example]
Transformation = Callable[[Example],
                          Optional[Example]]

annotate(train: Examples, fn: Transformation)
    -> Examples
\end{lstlisting}
\end{samepage}

Akin to a specialized \texttt{map}, the \texttt{annotate} primitive accepts a user-defined transformation \texttt{fn} and applies it over a list of training examples. Whenever \texttt{fn} returns an example (rather than \texttt{None}), \texttt{annotate} caches the intermediate predictions (i.e., the generated queries and retrieved passages). These predictions serve as successful demonstrations for the pipeline transformations. In simple uses, \texttt{fn} may attempt to answer the example ``zero-shot'' one or more times. This is typically done by invoking the \search and \predict stages of the program. When an answer is produced, if \texttt{fn} assesses it as correct, it returns a populated example in which the intermediate predictions are present.

\vspace{-2mm}
\paragraph{Case Study}

The snippet below defines the function \texttt{multihop\_demonstrate}, called in Line~3 of \texttt{multihop\_program}, and illustrates the usage of \texttt{annotate}.

\begin{samepage}
\begin{lstlisting}[language=Python,breaklines=true]
from dsp import sample, annotate

def attempt_example(d: Example):
   d = d.copy(demos=[])
   d = multihop_search(d)
   d = multihop_predict(d)
   return d if d.pred == d.answer else None

def multihop_demonstrate(x: Example):
   demos = annotate(x.train, attempt_example)
   return Example(x, demos=demos)
\end{lstlisting}
\end{samepage}

In Line~10, \texttt{multihop\_demonstrate} invokes \texttt{annotate}, which bootstraps missing fields in training examples by caching annotations from \texttt{attempt\_example}. The transformation \texttt{attempt\_example} takes a training example \texttt{d} and attempts to answer it in a zero-shot fashion: it creates a copy of \texttt{d} with no demonstrations (Line~4; i.e., zero-shot) and invokes the multi-hop search and predict pipeline (Lines~5 and~6). Each transformation returns an updated version of \texttt{d} with additional fields populated. If the pipeline answers correctly (Line~7), the updated \texttt{d} is returned.

Figure~\ref{fig:example} illustrates this behavior. \demonstrate transforms a training question--answer pair to a fully-populated demonstration, including fields such as \texttt{hop1} and \texttt{hop2} (i.e., queries for multi-hop search) as well as \texttt{psg1} and \texttt{psg2}. When the \LM{} is later invoked to conduct a transformation, say, generating a ``second-hop'' query during \search, the \texttt{psg1} field serves as context and the \texttt{hop2} field serves as a label for this particular training example.

\vspace{-2mm}
\paragraph{Discussion}

This simple case study illustrates the power of composition in the \DSP abstraction. Because the pipeline is a well-defined program in which transformations communicate by passing text attached to \texttt{Example}s, a simple map-and-filter strategy can leverage the \LM{} and \RM{} to bootstrap annotations for a full \textit{pipeline} from end-task labels. This is an extensible strategy, but even in its simplest form it generalizes the approaches explored recently by \citet{zelikman2022star}, \citet{wei2022chain}, \citet{zhang2022automatic}, and \citet{huang2022large} in which an \LM{} self-generates chain-of-thought rationales for an individual prompt.

By bootstrapping pipelines, \demonstrate makes it easy to explore complex strategies in \search and \predict without writing examples for every transformation. This includes strategies that are challenging to explore without custom annotations in traditional retrieval-augmented NLP. For instance, \citet{khattab2021baleen} introduces a pipeline for multi-hop reasoning that is trained with weak supervision, extending work by \citet{lee2019latent} and \citet{khattab2021relevance}. In it, the target 3 or 4 passages that need to retrieved must be labeled but the system discovers the best \textit{order} of ``hops'' automatically.

In contrast, \DSP allows us to build complex pipelines without labels for intermediate steps, because we can compose programs out of small transformations. If \LM{} and \RM{} can accurately process such transformations ``zero-shot'' (i.e., without demonstrations) on at least one or two examples, these examples can be discovered with end-task labels and used as demonstrations.

To draw on our earlier analogy with DNN frameworks like PyTorch, \demonstrate aims to replace the function of backpropagation in extensible ways by simulating the behavior of the program (corresponding to a ``forward'' pass) and programmatically learning from errors. In doing this with frozen models and with only end-task labels, \demonstrate introduces a high degree of modularity. In particular, without hand-labeling intermediate transformations, developers may swap the training domain, update the training examples, or modify the program's strategy, and use \texttt{annotate} to automatically populate all of the intermediate fields for demonstrations.

\vspace{-2mm}
\paragraph{Selecting Demonstrations} It is not always possible to fit all of the training examples in the context window of the \LM{}. \DSP provides three primitives for selecting a subset of training examples, namely, \texttt{sample}, \texttt{knn}, and \texttt{crossval}.

\begin{samepage}

\begin{lstlisting}[language=Python,breaklines=true,showstringspaces=false]
sample(train: Examples, k: int)
    -> Examples

knn(train: Examples, cast: Callable[[Example], str])
    -> fn(example: Example, k: int) # currying
    -> Examples

crossval(train: Examples, n: int, k: int)
    -> fn(evaluate: Transformation)
    -> Examples
\end{lstlisting}
\end{samepage}

As a baseline choice, $k$ demonstrations can be randomly sampled from \texttt{train} using the \texttt{sample} primitive, an approach used by \citet{brown2020language} and much subsequent work. We can also leverage the \RM{}'s representations and select from the training set the $k$ nearest neighbors to the input text, a strategy explored by \citet{liu2021makes}. Another strategy is to apply cross-validation to select among a number of sampled sets of demonstrations \cite{perez2021true}. For example, given $|\texttt{train}| = 100$ training examples, \texttt{crossval} would select $n$ subsets of $k = 5$ examples each, and return the set with which a transformation \texttt{evaluate} performs best on the remaining $95$ examples.

\vspace{-2mm}
\paragraph{Compositions \& Extensions}

By manipulating demonstrations and higher-order transformations, these simple selection and bootstrapping primitives can be combined to conduct larger novel strategies. If the training set is very large (e.g., $|\texttt{train}| = 100,000$), we can conduct \texttt{knn} to find the nearest $k=16$ examples and only \texttt{annotate} these, arriving at a system that learns incrementally in real-time. If the training set is moderately large (e.g., $|\texttt{train}| = 1000$), we can conduct \texttt{crossval} and cache the performance of all prompts it evaluates on each training example. At test time, we can use \texttt{knn} to find $k=50$ similar examples to the test input and select the prompt that performs best on these $k$ examples, producing an adaptive system that is informed by the quality of its pipeline on different types of examples.

\subsection{\search}
\label{sec:dsp:search}

The \search stage gathers passages to support transformations conducted by the \LM{}. We assume a large knowledge corpus---e.g., a snippet of Web, Wikipedia, or arXiv---that is divided into text \textit{passages}.
Providing passages to the \LM{} facilitates factual responses, enables updating the knowledge store without retraining, and presents a transparency contract: when in doubt, users can check whether the system has faithfully used a reliable source in making a prediction.

In the simplest scenarios, \search can directly query the \RM{}, requesting the top-$k$ passages (from a pre-defined index) that match an input question. This baseline instantiation of \search simulates retrieval in most open-domain question answering systems, which implement a ``retrieve-then-read'' pipeline, like \citet{lee2019latent}, \citet{khattab2021relevance}, \citet{lazaridou2022internet}, and many others.

\begin{samepage}

\begin{lstlisting}[language=Python,breaklines=true]
from dsp import retrieve

def simple_search(x):
    passages = retrieve(query=x.question, k=2)
    return passages
\end{lstlisting}
\end{samepage}

\vspace{-2mm}
\paragraph{\search Strategies}

In many scenarios, the complexity of the task demands more sophisticated \search strategies that empower the \RM{} to find relevant passages. Our running example (Figure~\ref{fig:example}) is one such scenario, in which we suspect examples are likely to require \textit{multi-hop} reasoning in particular. Other settings, for instance, pose conversational challenges, in which the information need expressed by a user can only be resolved by taking into account previous turns in the conversation, or demand more extensive planning~\cite{zhong2022romqa}.

In the retrieval-augmented NLP literature, multi-hop search~\cite{xiong2020answering,khattab2021baleen} and conversational search~\cite{del2021question,raposo2022question} pipelines have received much attention. These systems are typically fine-tuned with many hand-labeled query ``rewrites''~\cite{anantha2020open}, ``decompositions''~\cite{geva2021did,min2019multi}, or target hops~\cite{yang2018hotpotqa,jiang2020hover}. %
Supported with automatic annotations from \demonstrate, the \search stage allows us to simulate many such strategies and many others
in terms of passing queries, passages, and demonstrations between the \RM{} and \LM{}. More importantly, \search facilitates our vision of advanced strategies in which the \LM{} and \RM{} cooperate to incrementally plan a research path for which the \RM{} gathers information and the \LM{} identifies next steps.

\paragraph{Case Study} Let us build on our running multi-hop example as a case study. We can define \texttt{multihop\_search\_v2} (Line~4 in our core program), a slightly more advanced version of the \search transformation from Figure~\ref{fig:example}. This transformation simulates the iterative retrieval component of fine-tuned retrieval-augmented systems like IRRR~\cite{qi2020retrieve}, which reads a retrieved passage in every hop and generates a search query (or a termination condition to stop hopping), and Baleen~\cite{khattab2021baleen}, which summarizes the information from many passages in each hop for inclusion in subsequent hops.

\begin{samepage}

\begin{lstlisting}[language=Python]
from dsp import generate

def multihop_search_v2(x, max_hops=3):
  x.hops = []
  
  for hop in range(max_hops):
    summary, query = generate(hop_template)(x)
    x.hops.append((summary, query))
    
    if query == 'N/A': break
    
    passages = retrieve(query, k=5)
    x.context = [summary] + passages
    
  return x
\end{lstlisting}
\end{samepage}

In \texttt{multihop\_search\_v2}, Line~7 calls the \texttt{generate} primitive, which invokes the \LM{} to produce a query for each retrieval hop. The \LM{} is conditioned on a prompt that is prepared using the \texttt{hop\_template} template. (We discuss prompt templates and the \texttt{generate} primitive in~\secref{sec:dsp:predict}.) Here, this template may be designed to generate a prompt that has the following format (e.g., for the second hop).

\begin{samepage}
\begin{lstlisting}[breaklines=true]
My task is to write a simple query that gathers information for answering a complex question. I write N/A if the context contains all information required.

{Task demonstrations from x.demos, if any}

Context: {x.context}
Question: {x.question}
Summary: Let's summarize the above context. __{summary}__
Search Query: __{query}__
\end{lstlisting}
\end{samepage}

As shown, the \LM{} is instructed to read the context retrieved in earlier hops and a complex question. It is then prompted to write: \textbf{(i)} a summary of the supplied context and \textbf{(ii)} a search query that gathers information for answering that question. The generated text will be extracted and assigned to the \texttt{summary} and \texttt{query} variables in (\texttt{multihop\_search\_v2}; Line~7). On Line~10, we terminate the hops if the query is ``N/A''. Otherwise, Line~12 retrieves $k=5$ passages using the query and Line~13 assigns the \texttt{context} for the subsequent hop (or for \predict), setting that to include the \texttt{summary} of all previous hops as well as the passages retrieved in the final hop so far.

\paragraph{Comparison with self-ask} It may be instructive to contrast this multi-hop \DSP program with the recent ``self-ask''~\cite{press2022measuring} prompting technique, which we compare against in~\secref{sec:eval}. Self-ask can be thought of as a simple instantiation of \DSP's \search stage. In it, the \LM{} asks one or more ``follow-up questions'', which are intercepted and sent to a search engine. The search engine's answers are concatenated into the prompt and are used to answer the question. This is essentially a simplified simulation of IRRR~\cite{qi2020retrieve}.

As a general framework, \DSP can express ideas like self-ask and many other, more sophisticated pipelines as we discuss in the present section. More importantly, \DSP offers a number of intrinsic advantages that lead to large empirical gains: 80\%--290\% over self-ask. For instance, \DSP programs are deeply modular, which among other things means that \DSP programs will annotate and construct their own demonstrations. Thus, they can be developed without labeling any of the intermediate transformations (e.g., the queries generated). In addition, the \LM{} prompts constructed by \DSP get automatically updated to align with the training data and retrieval corpus provided. In contrast, approaches like self-ask rely on a hand-written prompt with hard-coded examples.

Moreover, \DSP assigns the control flow to an explicit program and facilitates design patterns that invoke the \LM{} (or \RM{}) to conduct small transformations. This allows us to build steps that are dedicated to generating one or more retrieval queries, summarizing multiple passages per hop, and answering questions. These steps are individually simpler than the self-ask prompt, yet our multi-hop \DSP program deliberately composes them to build richer pipelines that are thus more reliable. In contrast, self-ask delegates the control flow to the \LM{} completions, maintaining state within the prompt itself and intercepting follow-up questions to conduct search. We find that this paradigm leads to a ``self-distraction'' problem (\secref{sec:eval:datasets}) that \DSP programs avoid.

\paragraph{Fusing Retrieval Results} For improved recall and robustness, we can also \textit{fuse} the retrieval across multiple generated queries. Fusion has a long history in information retrieval~\cite{fox1994combination,xue2013modeling,kurland2018fusion} and sequentially processing multiple queries was explored recently by \citet{gao2022attributed} for retroactively attributing text generated by LMs to citations. Inspired by these, we include a \texttt{fused\_retrieval} primitive to \DSP to offer a versatile mechanism for interacting with frozen retrievers. It accepts an optional fusion function that maps multiple retrieval lists into one. By default, \DSP uses a variant of CombSUM~\cite{fox1994combination}, assigning each passage the sum of its probabilities across retrieval lists.

To illustrate, the modification below generates $n=10$ queries for the transformation \texttt{multihop\_search\_v2}.

\begin{samepage}
\begin{lstlisting}[numbers=none,language=Python,breaklines=true]
c = generate(hop_template, n=10)(x)
passages = fused_retrieval(c.queries, k=5)
summary = c.summaries[0] # highest-scoring summary
\end{lstlisting}
\end{samepage}

\paragraph{Compositions \& Extensions} To illustrate a simple composition, we can equip a chatbot with the capacity for conversational multi-hop search by combining a query rewriting step, which produces a query that encompasses all of the relevant conversational context, with the multi-hop transformation, as follows.

\begin{samepage}
\begin{lstlisting}[language=Python,breaklines=true]
def conversational_multihop_search(x):
    x.question = generate(conv_rewriting_template)(x)
    return multihop_search_v2(x)
\end{lstlisting}
\end{samepage}

Similar approaches can be used for correcting spelling mistakes or implementing pseudo-relevance feedback~\cite{cao2008selecting,wang2022colbert}, in which retrieved passages are used to inform a better search query, though this has not been attempted with pretrained LMs to our knowledge.

\subsection{\predict}
\label{sec:dsp:predict}

The \predict stage generates the system output using demonstrations (e.g., in \texttt{x.demos}) and passages (e.g., in \texttt{x.context}). \predict tackles the challenges of reliably solving the downstream task, which integrates much of the work on in-context learning in general. Within \DSP, it also has the more specialized function of systematically aggregating information across a large number of demonstrations, passages, and candidate predictions.

\paragraph{Generating Candidates} Generally, \predict has to produce one or more candidate predictions for the end-task. To this end, the basic primitive in \predict is \texttt{generate}, which accepts a \texttt{Template} and (via currying) an \texttt{Example} and queries the \LM{} to produce one or more completions, as explored earlier in~\secref{sec:dsp:search}. A corresponding primitive that uses the \RM{} in this stage is \texttt{rank}, which accepts a query and one or more passages and returns their relevance scores.

\begin{samepage}

\begin{lstlisting}[language=Python,breaklines=true]
Template  # template: an object that can produce prompts and parse completions

generate(template: Template)
    -> fn(example: Example)
    -> Completions  # object with keys to access extracted preds and scores

rank(query: str, passages: List[str])
    -> List[float]  # object with keys to access passage texts and scores
\end{lstlisting}
\end{samepage}

A \texttt{Template} is an object that can produce prompts, that is, map an \texttt{Example} to a string, and extract fields out of completions. For instance, we can map an example \texttt{x} that has a question and retrieved passages to the following prompt:

\begin{samepage}

\begin{lstlisting}[breaklines=true]
My task is to answer questions using Web documents.

{Task demonstrations from x.demos, if any}

Context: {x.passage}
Question: {x.question}
Rationale: Let's think step by step. __{rationale}__
Answer: __{answer}__
\end{lstlisting}
\end{samepage}

As this illustrates, the \LM{} will be asked to generate a chain-of-thought rationale (CoT; \citealt{wei2022chain}; \citealt{kojima2022large}) and an answer, and the generated text will be extracted back into the \texttt{rationale} and \texttt{answer} keys of each completion.

Each invocation to the \LM{} can sample multiple candidate predictions. Selecting a ``best'' prediction is the subject of much work on decoding \cite{wiher2022decoding,li2022contrastive}, but a frozen and general-purpose \LM{} may not support custom modifications to decoding. Within these constraints, we present several high-level strategies for selecting predictions and aggregating information in \DSP via the \LM{} and \RM{}.

\paragraph{Selecting Predictions} Among multiple candidates, we can simply extract the most popular prediction. When a CoT is used to arrive at the answer, this is the self-consistency method of \citet{wang2022self}, which seeks to identify predictions at which multiple distinct rationales arrive.

\begin{samepage}
\begin{lstlisting}[language=Python,breaklines=true]
from dsp import generate, majority

def multihop_predict(x):
   candidates = generate(template_qa)(x)
   return x.copy(answer=majority(candidates).answer)
\end{lstlisting}
\end{samepage}

\DSP generalizes this in two ways. First, we can sample multiple ``pipelines of transformations'' (PoT) within the program, rather than locally with ``chains of thought'' (CoT) in one transformation. These chains may even invoke different paths in the program, as illustrated below.

\vspace{8mm}
\begin{samepage}
\begin{lstlisting}[language=Python,breaklines=true]
from dsp import branch

def pipeline(x):
  return multihop_predict(multihop_search_v2(x))

def PoT_program(question: str) -> str:
  x = Example(question=question, train=train)
  x = multihop_demonstrate(x)
  
  candidates = branch(pipeline, n=5, t=0.7)(x)
  return x.copy(answer=majority(candidates).answer)
\end{lstlisting}
\end{samepage}

In the snippet above, Line~10 invokes the primitive \texttt{branch} which samples $n$ different PoTs with a high temperature (e.g., $t=0.7$) and accumulates their intermediate and final predictions. In this example, our pipeline invokes \texttt{multihop\_search\_v2} (\secref{sec:dsp:search}), which applies a variable number of retrieval hops depending on the questions generated, before doing \predict. That is, \texttt{PoT\_program} potentially invokes multiple distinct paths in the program (i.e., with different multi-hop queries and number of hops in each) across branches. It then selects the \texttt{majority} answer overall.

\DSP generalizes self-consistency in a second way. When sampling our CoTs or PoTs provides multiple candidates, we can select the top-$k$ (e.g., top-4) predictions and then \textit{compare} them directly. For instance, we may prompt the \LM{} to compare these choices as MCQ candidates, a transformation for which \demonstrate can automatically prepare exemplars. This effectively simulates the LM recursion of \citet{levine2022standing}, though unlike their approach it does not require a large training set or updating any (prompt-tuning) weights. One such implementation is illustrated in \texttt{openqa\_predict} below.

\begin{samepage}
\begin{lstlisting}[language=Python,breaklines=true]
def openqa_predict(x):
   preds = generate(template_qa, n=20)(x).answers
   x.choices = most_common(preds, k=4)
   
   queries = [f"{x.question} {c}"
              for c in x.choices]
              
   x.passages = fused_retrieval(queries)
   x.answer = generate(TemplateMCQ)(x).answer
   return x
\end{lstlisting}
\end{samepage}

As an alternative comparison approach, we can invoke the \RM{} via \texttt{rank} to find the prediction that is most grounded in a retrieved contexts (i.e., most similar to the concatenation of the retrieved passages) or, given an \RM{} that can score completions~\cite{krishna2022rankgen}, simply the prediction that has the highest score given the prompt.

\vspace{-2mm}
\paragraph{Aggregating Information}  When only a few demonstrations or passages are selected, we can simply concatenate them all into the prompt. For instance, GPT-3.5 \texttt{text-davinci-002} has a context window of 4097 tokens, which we find to be reasonably large for accommodating several (e.g., 3--5) demonstrations, which individually include their own passages and rationales.

To deal with a larger number of demonstrations or passages, we can \texttt{branch} in parallel to process individual subsets of the passages or demonstrations and then aggregate the individual answers using one of the scoring methods presented earlier. Indeed, \citet{lewis2020retrieval} and \citet{lazaridou2022internet} have explored marginalization as a way to combine scores across passages and \citet{le2022few} ensemble prompts across demonstrations, which can be expressed in this way.

An alternative aggregation strategy is to accumulate information across passages sequentially, rather than independently. This is effectively how our multi-hop approach works (\secref{sec:dsp:search}). Such a strategy has also been employed recently by \citet{gao2022attributed} for retroactively attributing text generated by LMs to citations. They generate many queries but instead of fusion (\secref{sec:dsp:search}), they run their pipeline on each query and use its outputs to alter the input to subsequent queries.\footnote{Though most of the functionality in this section is implemented, the primitives \texttt{branch}, \texttt{knn}, and \texttt{crossval} are currently work-in-progress.}

\section{Evaluation}
\label{sec:eval}

We now consider how to implement DSP programs for three diverse knowledge-intensive NLP tasks: open-domain question answering (QA), multi-hop QA, and conversational QA. All of these tasks are ``open-domain'', in the sense that systems are given a short question or participate in a multi-turn conversation without being granted access to context that answers these questions.

We build and evaluate intuitive compositions of the functions explored in \secref{section:dsp} for each task. We show that, despite low development effort, the resulting \DSP programs exhibit strong quality and deliver considerable empirical gains over vanilla in-context learning and a standard retrieve-then-read pipeline with in-context learning.

\subsection{Evaluation Methodology}

In this report, we consider one \textit{development dataset} for each of the tasks we consider, namely, the open-domain version of SQuAD~\cite{rajpurkar2016squad,lee2019latent}, the multi-hop HotPotQA~\cite{yang2018hotpotqa} dataset in the open-domain ``fullwiki'' setting, and the conversational question answering QReCC~\cite{anantha2020open,vakulenko-etal-2022-scai} dataset, which we used for developing the \DSP abstractions. We report the validation set accuracy on all three datasets and discuss them in detail \secref{sec:eval:datasets}.

Unless otherwise stated, systems are given access to 16-shot training examples, that is, each DSP program can use (up to) 16 questions---or conversations, where applicable---randomly sampled from the respective training set. We subsample the validation and test sets to 1000 questions (or 400 conversations, where applicable) and report average quality across five seeds where each seed fixes a single $k$-shot training set of examples. To control the language model API spending budget, each seed processes one fifth of the evaluation examples (e.g., 200 questions per seed, for a total of 1000 unique questions).

We also dedicate held-out \textit{test datasets} (e.g., Open-NaturalQuestions; \citealt{kwiatkowski2019natural}) and \textit{test tasks} (e.g., claim verification, as in FEVER; \citealt{thorne2018fever}) that we only use for evaluating pre-defined DSP programs rather than development. We will include these results in a future version of this report.

\subsection{Pretrained Modules}

\paragraph{\RM{}} We use ColBERTv2~\cite{santhanam-etal-2022-colbertv2}, a state-of-the-art retriever based on late interaction~\cite{khattab2020colbert}. We choose ColBERTv2 for its highly effective zero-shot search quality and efficient search~\cite{santhanam2022plaid}. %
However, our \DSP{} programs are agnostic to how the retriever represents examples or scores passages, so essentially any retriever can be used.

In addition, by making retrieval a first-class construct, \DSP allows us to change or update the search index over time. We simulate this in our experiments by aligning each of our datasets with the nearest Wikipedia corpus among the Dec 2016 Wikipedia dump from~\citealt{chen2017reading}, the Nov 2017 Wikipedia ``abstracts'' dump from ~\citealt{yang2018hotpotqa}, and the Dec 2018 Wikipedia dump from~\citealt{karpukhin2020dense}.

\paragraph{\LM{}} We use the GPT-3.5 (\texttt{text-davinci-002}; \citealt{brown2020language}; \citealt{ouyang2022training}) language model. Unless otherwise stated, we use greedy decoding when generating $n=1$ prediction. We sample with temperature $t=0.7$ when $n>1$, like related work~\cite{wang2022self}. %

\subsection{Baselines}
\label{sec:eval:baselines}

\paragraph{Vanilla LM} The vanilla LM baselines represent the few-shot in-context learning paradigm used by \citet{brown2020language}. The open-domain QA and multi-hop QA baselines randomly sample 16 demonstrations (i.e., all of the examples available to each program in our evaluation) from the training set and do not augment these demonstrations with evidence. Similarly, the conversational QA baseline samples four conversations. The vanilla baselines do not  search for passages relevant to the input query.

\begin{lstlisting}[numbers=left,language=Python,breaklines=true]
def vanilla_LM_QA(question: str) -> str:
    demos = sample(train, k=16)
    x = Example(question=question, demos=demos)
    return generate(qa_template)(x).pred
\end{lstlisting}

\paragraph{Retrieve-then-Read} The ``retrieve-then-read'' baselines use the \RM{} to support each example with a potentially relevant passage before submitting the prompt to the \LM{}. This emulates the pipelines used by state-of-the-art open-domain question answering systems~\cite{khattab2021relevance,izacard2020leveraging,hofstatter2022fid}. In conversational QA, we concatenate the first turn and the final question, an approach that we found to perform much better than simply using the final turn. For multi-hop QA, we retrieve and concatenate two passages per question.

\begin{lstlisting}[numbers=left,language=Python,breaklines=true]
def retrieve_then_read_QA(question: str) -> str:
    demos = sample(train, k=16)
    passages = retrieve(question, k=1)
    x = Example(question=question,
                passages=passages,
                demos=demos)
    return generate(qa_template)(x).pred
\end{lstlisting}

\paragraph{Self-ask} We also compare against self-ask~\cite{press2022measuring}, a contemporaneous pipeline that can be thought of as a specific instantiation of \DSP's \search stage followed by a simple \predict step. For direct comparison with our methods, we modify the self-ask control flow to query the same ColBERTv2 index used in our \DSP experiments instead of Google Search. We evaluate two configurations of self-ask. The first uses the original self-ask prompt template, which contains four hand-written demonstrations. In the second configuration, we modify the prompt template to apply a number of changes that we find are empirically useful for HotPotQA.\footnote{In particular: \textbf{(i)} use ColBERTv2-style passages in the hand-crafted demonstrations of self-ask (i.e., instead of the original Google-style snippets), \textbf{(ii)} concatenate 16-shot training examples from the task (i.e., question--answer pairs) as a prefix of the prompt, \textbf{(iii)} ask the model to generate a short intermediate answer per retrieval step, and \textbf{(iv)} explicitly ask the model to generate a follow-up ``search query'' at each step. We found the final item to be important because self-ask's default prompt often produces follow-up questions that are not self-contained (e.g., ``what is the name of the national park?'', which is not an informative search query). We also fix the casing in the prompt to be consistent.}

\newcommand{\phmark}{\phantom{$^\mathparagraph$}}
\begin{table*}[tp]
\centering
\small
\caption{Development results comparing a task-aware DSP program against baseline vanilla LM and retrieve-then-read LM as well as recent and contemporaneous in-context learning approaches with and without retrieval. All of our runs use GPT-3.5 and our retrieval-based rows use ColBERTv2. The results marked with $^\mathparagraph$ are collected from related work as of mid-December 2022, and attributed to their individual sources in the main text. As we discuss in the main text, the marked results are not generally apples-to-apples comparisons, since they span a variety of evaluation settings. Nonetheless, we report them here as qualitative reference points.
}
\vspace{2mm}
\begin{tabular}{ l @{\hspace{32pt}} cc @{\hspace{42pt}}  cc @{\hspace{42pt}} cc @{\hspace{8pt}}}
\toprule
& \multicolumn{2}{c@{\hspace{42pt}}}{\textbf{Open-SQuAD}}
& \multicolumn{2}{c@{\hspace{42pt}}}{\textbf{HotPotQA}}
& \multicolumn{2}{c@{\hspace{42pt}}}{\textbf{QReCC}} \\
& EM & F1 & EM & F1 & F1 & nF1 \\
\midrule
\textbf{Vanilla LM}            & 16.2\phmark & 25.6 & 28.3\phmark & 36.4 & 29.8 & 18.4 \\
\textbf{No-retrieval LM SoTA} & 20.2$^\mathparagraph$ & -- & 33.8$^\mathparagraph$ & 44.6$^\mathparagraph$ & -- & -- \\
\midrule
\textbf{Retrieve-then-Read}  & 33.8\phmark & 46.1 & 36.9\phmark & 46.1 & 31.6 & 22.2 \\
\textbf{Self-ask} w/ ColBERTv2 Search  & \phantom{0}9.3\phmark & 17.2 & 25.2\phmark & 33.2 & -- & -- \\
\textbf{\hspace{3mm} + Refined Prompt}  &  \phantom{0}9.0\phmark & 15.7 & 28.6\phmark & 37.3 & -- & -- \\
\textbf{Retrieval-augmented LM SoTA} & 34.0$^\mathparagraph$ & -- &  35.1$^\mathparagraph$ & -- & -- & -- \\
\midrule
\textbf{Task-aware DSP Program} & \textbf{36.6}\phmark & \textbf{49.0} & \textbf{51.4}\phmark & \textbf{62.9} & \textbf{35.0} &  \textbf{25.3} \\
\bottomrule
\end{tabular}%
\label{table:dev-results}
\end{table*}

\subsection{Proposed DSP Programs}

We build on transformations presented in~\secref{sec:dsp}. Our programs for all three tasks have the following structure, illustrated for open-domain QA.

\vspace{3mm}
\begin{lstlisting}[numbers=left,language=Python,breaklines=true]
def openqa_program(question: str) -> str:
   x = Example(question=question, train=train)
   x = openqa_demonstrate(x)
   x = openqa_search(x)
   x = openqa_predict(x)
   return x.answer
\end{lstlisting}

The exception is that the conversational QA program, \texttt{convqa\_program}, accepts \texttt{turns} (i.e., a list of strings, representing the conversational history) instead of a single \texttt{question}. Unless otherwise stated, our programs default to greedy decoding during the \demonstrate stage. %

For \search, our open-domain QA program uses the question directly for retrieving $k=7$ passages and concatenates these passages into our QA prompt with CoT. For \predict, it generates $n=20$ reasoning chains and uses self-consistency (SC; \citealt{wang2022self}) to select its final prediction. For \demonstrate, our open-domain QA program uses the following approach, slightly simplified for presentation. In it, the parameter $k=3$ passed to \texttt{annotate} requests annotating only three demonstrations, which will then be used in the prompts.

\begin{lstlisting}[numbers=left,language=Python,breaklines=true]
def openqa_demonstrate(x: Example) -> Example:
    demos = sample(x.train, k=16)

    def openqa_attempt(d: Example) -> Example:
        d.demos = all_but(demos, d)  # all (raw) examples different from d
        
        d = openqa_search(d, k=2)
        if not passage_match(d): return None  # skip examples where search fails
        
        d = openqa_predict(d, sc=False)
        if not answer_match(d): return None  # skip examples where predict fails
        
        return d

    x.demos = annotate(demos, openqa_attempt, k=3)
    return x
\end{lstlisting}

Our multi-hop program adopts a very similar approach for \demonstrate and \predict. For \search, it uses the approach described in~\secref{sec:dsp:search}, with the following adjustments. It uses result fusion across $n=10$ queries per hop and, among the $n$ predictions, uses the summary corresponding to the largest average log-probability. It uses a fixed number of hops for HotPotQA, i.e., two hops. In each prompt (i.e., each hop and QA), it concatenates the summaries of all previous hops (i.e., hop 1 onwards) and a total of $k=5$ passages divided between the hops (i.e., five passages from the first hop or two passages from the first and three from the second).

For conversational QA, we use a simple \predict which generates a response with greedy decoding, conditioned on all of the previous turns of the conversation and five retrieved passages. For \search, our conversational QA pipeline generates $n=10$ re-written queries (and also uses the simple query as the retrieve-and-read baseline; \secref{sec:eval:baselines}) and fuses them as in~\secref{sec:dsp:search}. We implement \demonstrate similar to \texttt{openqa\_demonstrate}, but sample only four examples (i.e., four conversational turns; instead of 16 questions as in open-domain QA) for demonstrating the task for the higher-order transformation \texttt{convqa\_attempt}, which is passed to \texttt{annotate} (not shown for brevity).

\begin{lstlisting}[numbers=left,language=Python,breaklines=true]
def convqa_attempt(d: Example) -> Example:
    d.demos = all_but(demos, d)  # all (raw) examples that don't intersect with the conversation of d

    d = convqa_search(d, k=2)
    if max(precision(d.answer, p) for p in d.passages) < .8: return None  # skip examples where search fails

    d = convqa_predict(d, n=20)
    if max(F1(c.pred, d.answer) for c in d.candidates) < .75: return None  # skip examples where predict fails out of n=20 attempts

    return d
\end{lstlisting}

\subsection{Development Datasets \& Results}
\label{sec:eval:datasets}

\paragraph{Open-SQuAD} We conduct the open-domain version of SQuAD over the Wikipedia 2016 corpus from \citet{chen2017reading}, as processed by \citet{khattab2021relevance}. We use the same train/validation/test splits as in \citet{karpukhin2020dense} and \citet{khattab2021relevance}.

Table~\ref{table:dev-results} reports the answer EM and F1. The task-aware \DSP program achieves 36.6\% EM, outperforming the vanilla LM baseline by 126\% EM relative gains. This indicates the importance of grounding the \LM{}'s predictions in retrieval, and it shows that state-of-the-art retrievers like ColBERTv2 have the capacity to do so off-the-shelf. The proposed \DSP program also achieves relative gains of 8\% in EM and 6\% in F1 over the retrieve-then-read pipeline, highlighting that non-trivial gains are possible by aggregating information across several retrieved passages as we do with self-consistency.

These in-context learning results are competitive with a number of popular fine-tuned systems. For instance, on the Open-SQuAD test set, DPR achieves 29.8\% EM, well below our 16-shot \DSP program. On the Open-SQuAD dev set, the powerful Fusion-in-Decoder~\cite{izacard2020leveraging} ``base'' approach achieves approximately 36\% (i.e., very similar quality to our system) when invoked with five retrieved passages. Nonetheless, with the default setting of reading 100 passages, their system reaches 48\% EM in this evaluation. This may indicate that similar gains are possible for our \DSP program if the \predict stage is made to aggregate information across many more passages.

For comparison, we also evaluate the self-ask pipeline, which achieves 9.3\% EM, suggesting that its fixed pipeline is ineffective outside its default multi-hop setting. Studying a few examples of its errors reveals that it often decomposes questions in tangential ways and answers these questions instead. We refer to this behavior of the \LM as ``self-distraction'', and we believe it adds evidence in favor of our design decisions in \DSP. To illustrate self-distraction, when self-ask is prompted with ``When does The Kidnapping of Edgardo Mortara take place?'', it asks ``What is The Kidnapping of Edgardo Mortara`` and then asks when it was published, a tangential question. Thus, self-ask answers ``1997'', instead of the time The Kidnapping of Edgardo Mortara takes place (1858).

For reference, Table~\ref{table:dev-results} also reports (as No-retrieval LM SoTA) the concurrent in-context learning results from \citet{si2022prompting} using \texttt{code-davinci-002}, who achieve 20.2\% EM without retrieval and 34.0\% EM with retrieval, albeit on a different sample and split of the SQuAD data. Overall, their approaches are very similar to the baselines we implement (vanilla LM and retrieve-then-read), though their retrieval-augmented approach retrieves (and concatenates into the prompt) 10 passages from a Wikipedia dump.

\paragraph{HotPotQA} We use the open-domain ``fullwiki'' setting of HotPotQA using its official Wikipedia 2017 ``abstracts'' corpus. The HotPotQA test set is hidden, so we reserve the official validation set for our testing. We sub-divide the training set into 90\%/10\% train/validation splits. In the training (and thus validation) split, we keep only examples marked as ``hard'' in the original dataset, which matches the designation of the official validation and test sets.

We report the final answer EM and F1 in Table~\ref{table:dev-results}. On HotPotQA, the task-aware \DSP program outperforms the baselines and existing work by very wide margins, exceeding the vanilla LM, the retrieve-then-read baseline, and the self-ask pipeline by 82\%, 39\%, and 80\%, respectively, in EM. This highlights the effectiveness of building up more sophisticated programs that coordinate the \LM{} and \RM{} for the \search step.

These results may be pegged against the evaluation on HotPotQA in a number of concurrent papers. We first compare with non-retrieval approaches, though our comparisons must be tentative due to variation in evaluation methodologies. \citet{si2022prompting} achieve 25.2\% EM with CoT prompting. With a ``recite-and-answer'' technique for PaLM-62B~\cite{chowdhery2022palm}, \citet{sun2022recitation} achieve 26.5\% EM. \citet{wang2022rationale} achieve 33.8\% EM and 44.6 F1 when applying a self-consistency prompt for PaLM-540B. Next, we compare with a contemporaneous retrieval-based approach: \citet{yao2022react} achieve 35.1\% EM using a system capable of searching using a Wikipedia API. All of these approaches trail our task-aware \DSP program, which achieves 51.4\% EM, by large margins.

\paragraph{QReCC} We use QReCC~\cite{anantha2020open} in an open-domain setting over Wikipedia 2018. QReCC does not have an official development set, so we sub-divide the training set into 90\%/10\% train/validation splits. For the first question in every conversation, we use the rewritten question as the original question often assumes access to a ground-truth document. We also filter low-quality examples from QReCC.\footnote{We remove conversations that have one or more empty ground-truth answers and conversations that have only one or two questions. We also find many conversations that include ``what other interesting facts are in this article?'', which conflict with the open-domain formulation and have no well-defined answer. Hence, we remove any conversation that includes the keywords ``other interesting'' or ``else'', which we found to be markers of low quality.}

We conduct the QReCC conversations in an auto-regressive manner. At turn $t > 1$ of a particular conversation, the system sees its own responses (i.e., not the ground truth responses) to previous turns of the conversation. We report the novel-F1 metric (nF1; \citealt{paranjape2021hindsight}), which computes the F1 overlap between the system response and the ground truth while discounting common stopwords and terms present in the question (or earlier questions). The results are shown in Table~\ref{table:dev-results}, and follow the same general pattern as SQuAD and HotPotQA.

\section{Conclusion}

For a long time, the dominant paradigm for building models in AI has centered around multiplication of tensor representations, and in the deep learning era this has given rise to highly modular (layer-wise) designs that allow for fast development and wide exploration. However, these design paradigms require extensive domain expertise, and even experts face substantial challenges when it comes to combining different pretrained components into larger systems.

The promise of in-context learning is that we can build complex systems from pretrained components using only natural language as the medium for giving systems instructions and, as we argue for, allowing components to communicate with each other. In this new paradigm, the building blocks are pretrained models and the core operations are natural language instructions and operations on natural language texts. If we can realize this potential, then we can broaden participation in AI system development, rapidly prototype systems for new domains, and maximize the value of specialized pretrained components.

In the current paper, we introduced the \demonstrate--\search--\predict (\DSP) framework for retrieval augmented in-context learning. \DSP{} consists of a number of simple, composable functions for implementing in-context learning systems as deliberate \textit{programs}---instead of end-task prompts---for solving knowledge intensive tasks. We implemented \DSP{} as a Python library and used it to write programs for Open-SQuAD, HotPotQA, and QReCC. These programs deliver substantial gains over previous in-context learning approaches. However, beyond any particular performance number, we argue that the central contribution of \DSP{} is in helping to reveal a very large space of conceptual possibilities for in-context learning in general.

\section*{Acknowledgements}

We thank Ashwin Paranjape, Amir Ziai, and Rick Battle for valuable discussions and feedback.
This work was partially supported by IBM as a founding member of the Stanford Institute for Human-Centered Artificial Intelligence (HAI). This research was supported in part by affiliate members and other supporters of the Stanford DAWN project---Ant Financial, Facebook, Google, and VMware---as well as Cisco, SAP, and the NSF under CAREER grant CNS-1651570. Any opinions, findings, and conclusions or recommendations expressed in this material are those of the authors and do not necessarily reflect the views of the National Science Foundation.
We thank Giuseppe Attanasio for his public \LaTeX{} GitHub-style Python code formatting gist.\footnote{\url{https://gist.github.com/g8a9/07c2be12ae02cfad4aa430d77dc940cb}} We also thank Riley Goodside for his public tips on formatting LM prompts (at \texttt{@goodside} on Twitter).

\bibliography{custom,anthology}
\bibliographystyle{icml2021}

\appendix

\end{document}